\documentclass{article}
\usepackage[utf8]{inputenc}

\usepackage{amsfonts,amsmath,graphicx,palatino,epsfig,latexsym,natbib, array, stfloats, float}

\usepackage[ruled,vlined]{algorithm2e}
\SetKwBlock{Repeat}{repeat}{}
\SetKwFor{RepeatTimes}{repeat}{times}{end}
\usepackage{authblk}

\title{Evolutionary Algorithms for solving Unconstrained, Constrained and Multi-objective Noisy Combinatorial Optimisation Problems}
\author[$\dagger$]{Aishwaryaprajna}
\author[$\dagger$$\star$]{Jonathan E. Rowe}

\affil[$\dagger$]{School of Computer Science, University of Birmingham, Birmingham, United Kingdom}
\affil[$\star$]{The Alan Turing Institute, London, United Kingdom}
\date{}

\begin{document}

\maketitle
\begin{abstract}
    We present an empirical study of a range of evolutionary algorithms applied to various noisy combinatorial optimisation problems. There are three sets of experiments. The first looks at several toy problems, such as \textsc{OneMax} and other linear problems. We find that UMDA and the Paired-Crossover Evolutionary Algorithm (PCEA) are the only ones able to cope robustly with noise, within a reasonable fixed time budget. In the second stage, UMDA and PCEA are then tested on more complex noisy problems: \textsc{SubsetSum}, \textsc{Knapsack} and \textsc{SetCover}. Both perform well under increasing levels of noise, with UMDA being the better of the two. In the third stage, we consider two noisy multi-objective problems (\textsc{Counting- OnesCountingZeros} and a multi-objective formulation of \textsc{SetCover}). We  compare several adaptations of UMDA  for multi-objective problems with the Simple Evolutionary Multi-objective Optimiser (SEMO) and NSGA-II.  We conclude that UMDA, and its variants, can be highly effective on a variety of noisy combinatorial optimisation, outperforming many other evolutionary algorithms.
\end{abstract}
\section{Introduction}
Realistic optimisation problems are often affected with noisy fitness measurements. The recent theoretical analyses of evolutionary algorithms (EAs) on noisy problems defined in discrete spaces are mostly focused on simple and classical benchmark problems, such as \textsc{OneMax}. Harder and more realistic combinatorial problems such as \textsc{Knapsack} or \textsc{SetCover} in presence of noisy fitness evaluations are not easily amenable to theoretical analysis. This paper discusses a thorough empirical analysis of an array of EAs on several simple and harder noisy combinatorial problems. This study attempts to identify which EAs to choose when solving realistic noisy  combinatorial problems. 

Noise may affect fitness evaluation in a number of ways. It may be considered as \textit{prior}, in which the search point is randomly tampered with, and fitness evaluation is performed on the noisy search point.  Alternatively,  \textit{posterior} noise (which will be the focus of our study) is where a random value gets added to the fitness of a search point. With more complex combinatorial problems (e.g. ones with constraints) the noise may enter in different ways (for example, in the evaluation of those constraints). 
 
An early theoretical result by \cite{droste2004analysis} examined the performance of the hill-climber $(1+1)$-EA on \textsc{OneMax} with prior noise. This was generalised to the $(\mu+\lambda)$-EA by \cite{Giesen:2016aa}, showing that populations can help in both prior and posterior noise. They show the $(1+1)$-EA, however, can only tolerate posterior Gaussian noise when the variance is very small (less than $1/(4 \log n)$). It has been recognised for a long time that the population size can affect the ability of an EA to handle noise~(\cite{goldberg1991genetic,rattray1998noisy}). A more recent theoretical study by \cite{dang2015efficient} shows that a low mutation rate enables a particular mutation-population algorithm to handle arbitrary posterior noise for the \textsc{OneMax} problem in polynomial time, although the bounds given are large. Similarly, the compact genetic algorithm (cGA) is shown to handle noise with (large) polynomial runtime (\cite{Friedrich:2015aa}). A better asymptotic runtime for \textsc{OneMax} with posterior Gaussian noise is proved for the Paired Crossover Evolutionary Algorithm (PCEA) which just uses crossover, and no mutation (\cite{PrugelBennett}).

Of course, it is possible to handle noise simply by re-sampling the fitness of a potential solution many times, and taking the average as an estimate of the true fitness. Suppose the noisy problem is defined by taking a deterministic fitness function and adding Gaussian noise with mean 0 and variance $\sigma^2$. There is a general result~\cite{AKIMOTO201542} that states if the runtime of a black box algorithm on a problem with no noise is $T$, then  $\sigma^2 \log T$ samples are required at each step leading to a runtime of $\sigma^ 2 T \log T$. In the case of \textsc{OneMax}, the most efficient algorithm~\cite{Anil:2009:BSE:1527125.1527135} has a runtime of $\Theta(n/\log n)$. Using this algorithm with resampling, gives a runtime for noisy \textsc{OneMax} of $\Theta(\sigma^2 n)$. By contrast, the PCEA algorithm~\cite{PrugelBennett}, when $\sigma^2 = n$, has a runtime of $O(n (\log n)^2)$ which is already faster than the resampling algorithm. It has been suggested by ~\cite{doerr2019resampling} that using the median rather than mean provides a better estimate when resampling, but this is only significant when the variance is small (less than a constant).

A recent study by ~\cite{rowe2019benefits} of a new Voting algorithm on \textsc{OneMax} shows a runtime of $O(n \log n)$, when the variance of the noise distribution is $\sigma^2 = O(n)$ and in $O(\sigma^2 \log n)$ when the noise variance is greater than this. This upper bound is the best proven runtime that we are aware of to date. Some empirical results show that the use of voting in population based algorithms (UMDA, PCEA and cGA) are effective for large population sizes.

In this paper, we are interested in whether any of the algorithms with polynomial theoretical runtimes for noisy \textsc{OneMax} would be capable of solving combinatorial problems with added noise in practice, when given a reasonable but fixed time budget\footnote{This paper is an extended version of \cite{aish1}.}. We proceed in three stages. First we will experimentally compare a collection of algorithms on noisy \textsc{OneMax} and noisy \textsc{Linear} problems, to see which can find solutions within a reasonable amount of time (to be defined below), bearing in mind that the asymptotic bounds for some of these algorithms, while polynomial, are actually very large. Second, we will take those algorithms which pass this first test, and see how well they handle noise in three combinatorial problems: \textsc{SubsetSum}, \textsc{Knapsack} and \textsc{SetCover}. We choose these, as they have a `packing' structure which might make them amenable to algorithms which can solve noisy \textsc{OneMax} efficiently. We generate random problem instances within the `easy' regime (so that the algorithms can be expected to solve them when there is no noise) and then empirically study how they degrade with added Gaussian noise. \par
In the last stage, we look at noisy multi-objective problems. Initially, we  analyse the performance of a collection of multi-objective algorithms on a toy multi-objective problem COCZ without and with high levels of noise and we attempted to identify which algorithms perform better. We study the simple hill-climber algorithm SEMO, the popular NSGA-II and some other algorithms designed on the basis of our previous experimental results. We compare our algorithms on the basis of the performance indicator, hypervolume, which provides an analysis of the spread of the non-dominated solutions found, in a reasonable time budget. We then formulate the noisy constrained \textsc{SetCover} problem as a multi-objective problem and we empirically analyse the performance of the better algorithms on this. 

It should be noted that in our empirical results, while error bars are not always shown, the Mann-Whitney test was used on all relevant comparisons, and results are significant at the 95\% level unless explicitly indicated.

{\bf Notation: } We use the convention that $[expr]$ equals 1 if $expr$ is true, and $0$ otherwise.

\section{Problem Definitions --- Noisy Single Objective Problems}
\label{sec:problem}
The problems studied in this paper are defined on a Boolean search space of bit strings of length $n$. Let $N(0, \sigma)$ denote a random number drawn from a normal distribution with mean zero, and standard deviation $\sigma$, which will be freshly generated at each fitness function evaluation.
\subsection{Unconstrained Single-objective noisy problems}

The first problem is \textsc{OneMax}, whose fitness function is defined as,
\[
	\textsc{OneMax}(x) = \sum_{i=1}^n x_i
\]
When the fitness evaluation is tampered with random noise, the fitness function becomes as follows,
\[
	\textsc{NoisyOneMax}(x) = \sum_{i = 1}^n x_i + N(0, \sigma)
\]

The \textsc{WeightedLinear} problem is defined with reference to $n$ positive weights $w_1, \ldots, w_n$ as follows,
\[
	\textsc{WeightedLinear}(x) = \sum_{i=1}^n x_i w_i
\]
with corresponding noisy variant,
\[
	\textsc{NoisyWeightedLinear}(x) = \sum_{i=1}^n x_i w_i + N(0, \sigma)
\]

In generating random problem instances, we draw the weights uniformly at random from the range $1, \ldots, 100$. Thus we avoid more extreme instances such as \textsc{BinVal} (in which $w_i = 2^{i-1}$ for each $i=1, \ldots, n$). The reason for this is that when the distribution of weights is highly skewed, the addition of noise is  irrelevant for those bits with very high weights, yet completely overwhelms bits with weights lower than the typical noise level. Thus most algorithms will find the more significant bits, and fail on the remainder.

The $\textsc{SubsetSum}$ problem is defined with reference to $n$ positive weights $w_1, \ldots, w_n$ and a target $\theta$,
\[
	\textsc{SubsetSum}(x) = | \theta - \sum_{i=1}^n x_i w_i|
\]
In presence of noisy fitness evaluations, the fitness function can be written as follows,
\[
	\textsc{NoisySubsetSum}(x) = | \theta - \sum_{i=1}^n x_i w_i| + N(0, \sigma)
\]

\textsc{SubsetSum} can be seen as a generalisation of the \textsc{WeightedLinear} problem (in which the target is $\theta = 0$). In our experiments, we generate instances by choosing weights uniformly at random from $1, \ldots, 100$. We take the target to be two-thirds of the sum of the weights (we have run experiments for other choices of $\theta$ and found that they do not significantly affect the empirical observations).

\subsection{Constrained single-objective noisy problems}

The $\textsc{Knapsack}$ problem is defined with respect to a set of positive weights $w_1, \ldots, w_n$, a capacity $C$ and positive profits $p_1, \ldots, p_n$ as follows,
\[
\textsc{Knapsack}(x) = 
\left\{
\begin{array}{lr}
\sum_{i=1}^n x_i p_i & \mbox{ if } \sum_{i=1}^n x_i w_i \leq C \\
C - \sum_{i=1}^n x_i w_i & \mbox{ otherwise }
\end{array}
\right.
\]
Random instances choose weights and profits uniformly from $1,\ldots,100$, and the capacity is two-thirds of the sum of the weights. We consider two noisy variants of the Knapsack problem. The first version simply considers posterior additive noise as before:
\[
\textsc{NoisyKnapsackV1}(x) = \textsc{Knapsack}(x) + N(0, \sigma)
\]

In the second version, the presence  of noise in the judgement with respect to the weights is considered,
\[
W_{\sigma}(x) = \sum_{i=1}^n x_i w_i + N(0, \sigma)
\]
If this (noisy) weight does not exceed the capacity, we then evaluate (noisily), the profit. Otherwise we return the excess weight:
\[
\textsc{NoisyKnapsackV2}(x) =
\]
\[
\left\{
\begin{array}{lr}
\sum_{i=1}^n x_i p_i  + N(0, \sigma) & \mbox{ if } W_{\sigma}(x) \leq C \\
C - W_{\sigma}(x) & \mbox{ otherwise }
\end{array}
\right.
\]
Note that noise is added to the weight just once, when the constraint is checked, and the same value used to report the fitness value, in the case the constraint is violated.

The \textsc{SetCover} problem finds a minimal covering of $m$ elements with a collection of sets from $n$ pre-defined subsets. A Boolean matrix $a_{ij}$ with $n$-rows and $m$-columns is used to define the $n$ subsets $c_1, \ldots, c_n$:
\[
a_{i, j} = [i \in c_j]
\]
The optimal collection of the sets would have the least number of the sets needed to cover all the $m$ elements. The \textsc{SetCover} problem has several real-world applications such as the airline crew scheduling problem.

The problem can be defined as a constrained single-objective one, as well as, a single-objective problem with a penalty term. The problem can also be defined as a multi-objective problem (discussed later).

The \textsc{ConstrainedSetCover} problem has a constraint that checks if the solution covers each of the $m$ elements. The optimal solution would have the least number of sets needed to cover all the $m$ elements. It is defined as follows,
\[
\textsc{ConstrainedSetCover}(x) =  \sum_{j=1}^n x_j \]
\[
\text{ subject to}  \sum_{j=1}^n x_j a_{ij} \geq 1 ,\text{   } i \in {1, \dots, m}
\]

For comparison-based algorithms, we always prefer feasible solutions instead of infeasible solutions. Two feasible solutions are compared by their fitness values, whereas two infeasible solutions by their constraint violations. The noisy version of the problem arises if the judgements regarding the number of elements uncovered and the number of the subsets required is noisy.  
\[
\textsc{NoisyConstrainedSetCover}(x) =  \sum_{j=1}^n x_j + N(0, \sigma) \]
\[
\text{ subject  to}  \sum_{j=1}^n x_j a_{ij} + N(0, \sigma)\geq 1 ,\text{   } i \in {1, \dots, m}
\]
The fitness function of \textsc{SetCover} problem can also be defined by including a penalty term such that, if elements are under-covered by the considered collection of sets, a huge penalty $\mu$ is incurred. 
\[
\textsc{PenaltySetCover}(x) = \]
\[
\sum_{j=1}^n x_j + \mu \sum_i \max \bigg\{ 0, \bigg(1-\sum_{j=1}^n a_{ij}x_j\bigg)\bigg\}
\]
This gives rise to a corresponding noisy variant:
\[
\textsc{NoisyPenaltySetCover}(x) = \]
\[
\sum_{j=1}^n x_j + \mu \sum_i \max \bigg\{ 0, \bigg(1-\sum_{j=1}^n a_{ij}x_j\bigg)\bigg\} + N(0, \sigma)
\]

\section{Algorithms Chosen For Noisy Single-objective optimisation}
\subsection{The (1+1)-EA}
The $(1+1)$-EA uses a bitwise mutation operator that produces an offspring by flipping each bit of the parent string independently with  probability $1/n$. This can be considered as a randomised or stochastic hill-climber which considers only one point in the search space at a time and proceeds by trying to find a point which has a superior function value. In each iteration, only one function evaluation takes place. The expected runtime of the $(1+1)$-EA solving the non-noisy \textsc{OneMax} is $O(n\log n)$. The runtime remains polynomial in the posterior Gaussian noise case for $\sigma^2 < 1/(4 \log n)$, so we do not expect this algorithm to cope with anything but the smallest noise levels~(\cite{Giesen:2016aa}).
\subsection{Mutation-Population Algorithm}
It has long been recognised that populations can help an EA handle noise. The paper by ~\cite{goldberg1991genetic} developed a population sizing equation and instigated the adoption of variance-based population sizing. ~\cite{rattray1998noisy} showed that in weak selection limit, effects of Gaussian noise could be overcome by an appropriate increase of the population size. More recently, a population-based, non-elitist EA was analysed by Dang \& Lehre to study how it optimises the noisy \textsc{OneMax} problem with uniform, Gaussian and exponential posterior noise distributions (~\cite{dang2015efficient, Dang2016}). They considered a recently developed fitness-level theorem for non-elitist populations to estimate the expected running time for the said problems in noisy environment. In case of additive Gaussian noise $N(0,\sigma^2)$ with mutation rate  $\frac{\chi}{n}=\frac{a}{3\sigma n}$  and population size $\lambda=b\sigma^2 \ln n$ (where $a$ and $b$ are constants), the considered algorithm optimizes the \textsc{OneMax} problem in expected time $O(\sigma^7 n \ln (n) \ln ({\ln{ (n)}}))$. Similar results were shown for uniform and exponential noise distributions. Note that this is potentially very large, when the noise is large --- in excess of $n^{4.5}$ when $\sigma = \sqrt{n}$, although of course this is an upper bound, and we do not know the constants. 

\subsection{Compact Genetic Algorithm (cGA)}

The compact GA (cGA) is an EDA, introduced by ~\cite{harik1999compact}. cGA is able to average out the noise and optimize the noisy \textsc{OneMax} problem in expected polynomial time, when the noise variance $\sigma^2$ is bounded by some polynomial in $n$, as suggested in ~\cite{Friedrich:2015aa}. The paper introduced the concept of graceful scaling in which the runtime of an algorithm scales polynomially with noise intensity, and suggested that cGA is capable of achieving this. It is also suggested that there is no threshold point in noise intensity at which the cGA algorithm begins to perform poorly (by which they mean having super-polynomial runtime). They proved that cGA is able to find the optimum of the noisy \textsc{OneMax} problem with Gaussian noise of variance $\sigma^2$ after $O(K \sigma^2 \sqrt n \log Kn)$ steps when $K=\omega(\sigma^2 \sqrt n \log n)$, with probability $1-o(1)$. Note that this upper bound is in excess of $n^3$ when $\sigma = \sqrt{n}$.

\subsection{Population Based Incremental Learning (PBIL)}
The algorithm PBIL, proposed by ~\cite{baluja1994population} in 1994, combines genetic algorithms and competitive learning for optimising a function. We have included this algorithm as it is in some ways similar to the cGA, so we might expect it to have similar performance. We are not aware of any theoretical analysis of this algorithm on noisy problems. The runtime of PBIL on \textsc{OneMax} is known to be $O(n^{3/2}\log{n})$, for suitable choice of $\lambda$~(\cite{7851024}).

\subsection{Univariate Marginal Distribution Algorithm (UMDA)}
The Univariate Marginal Distribution Algorithm (UMDA) proposed by ~\cite{muhlenbein1997equation} belongs to the EDA schema. In some ways, it is therefore similar to cGA and PBIL. However, it can also be viewed as generalising the \emph{genepool} crossover scheme, in which bits are shuffled across the whole population (within their respective string positions). We have included UMDA then, to see if its behaviour is more like cGA and PBIL on the one hand (which emphasise an evolving distribution over bit values), or like PCEA on the other (which emphasises crossover). The UMDA algorithm initialises a population of $\lambda$ solutions, and sorts the population according to the fitness evaluation of each candidate solution. The best $\mu$ members of the population are selected to calculate the sample distribution of bit values in each position. The next population is generated from this distribution. There are two variants of UMDA, depending on whether the probabilities are constrained to stay away from the extreme values of 0 and 1, or not. It is known that if the population size is large enough (that is, $\Omega(\sqrt{n}\log{n})$) then this handling of probabilities at the margins is not required (\cite{Witt:2017:UBR:3071178.3071216}). Since we will work with a large population (to match the PCEA algorithm described below), we will not employ margin handling, unless otherwise stated. In our experiments we will take $\mu = \lambda/2$. We are not aware of any theoretical results concerning UMDA on problems with posterior noise, but the runtime on \textsc{OneMax} is known to be $O(n \log n)$ for $\mu = \Theta(\sqrt{n}\log n)$ --- see ~\cite{Witt:2017:UBR:3071178.3071216}.

\subsection{Paired-Crossover EA (PCEA)}
Recently, the recombination operator has been suggested to be considerably beneficial in noisy evolutionary search. \cite{PrugelBennett} considered the problem of solving \textsc{OneMax} with noise of order $\sigma = \sqrt{n}$   and analysed the runtime of an evolutionary algorithm consisting only of selection and uniform crossover, the Paired-Crossover EA (PCEA). They show that if the population size is $c \sqrt{n} \log n$ then the required number of generations is $O\left(\sqrt{n} \log n\right)$, giving a runtime of $O(cn\left(\log n\right) ^2)$, with the probability of failure is $O(1/n^c)$. The proof in that paper can be generalised to the case of $\sigma \geq \sqrt{n}$, to give a runtime of $O(\sigma^2 \log n)$. It is not known what happens for lower levels of noise, though it is shown that in the absence of noise,  PCEA solves \textsc{OneMax} in $O(n (\log n)^2)$.

\section{Experiments --- Simple Noisy Problems}
\subsection{Noisy \textsc{OneMax}}
We investigate the performance of the algorithms described above, in solving the noisy \textsc{OneMax} problem. In the literature, some theoretical proofs exist for the expected runtime of specific algorithms on solving the noisy \textsc{OneMax} problem with additive posterior Gaussian noise (\cite{PrugelBennett, dang2015efficient, AKIMOTO201542, friedrich2017compact, lucas2017efficient, qian2018effectiveness, dang2018new, doerr2019resampling, doerr2020exponential}). We are interested in the algorithms' performances given a reasonable but fixed runtime budget across a wide range of noise levels, from $\sigma = 0$ up to $\sigma = \sqrt{n}$.  

To address the question of what constitutes a \emph{reasonable} budget, we compared the known theoretical results of our algorithms on noisy \textsc{OneMax}. PCEA has the lowest proven upper bound on its runtime, compared to the other algorithms for which results exist. We therefore allowed each algorithm to have twice the number of fitness evaluations that PCEA requires (on average) to find the optimum, as a reasonable budget. The function evaluation budgets calculated in this way are given in Table~\ref{tab:onemax_budgets}.\par
\begin{figure}
\begin{center}
\includegraphics[scale=0.42]{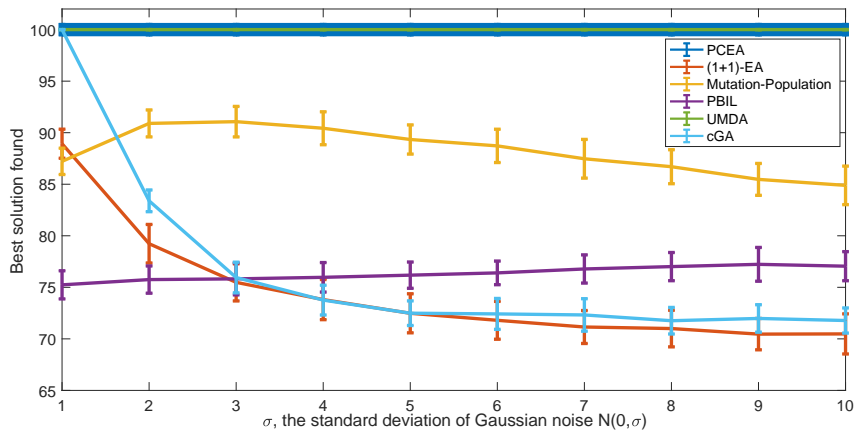}
\caption{Comparison of the algorithms while solving the noisy \textsc{OneMax} for different noise levels}
\label{fig:onemax_smallnoise}
\end{center}
\end{figure}

\begin{table}
    \centering
     
    \begin{tabular}{|c|c|c|c|c|c|}
    \hline
         $\sigma$ &  1 & 2 & 3 & 4 & 5  \\
         budget & 38392 & 41066 & 44477 & 50728 & 56851 \\
         \hline
         $\sigma$ &  6 & 7 & 8 & 9 & 10 \\
         budget & 64079 & 70736 & 79034 & 86078 & 93638\\
         \hline
    \end{tabular}
    \label{tab:onemax_budgets}
     \caption{Function evaluation budgets allowed for noisy \textsc{OneMax} experiments with different noise levels.}
\end{table}
\begin{figure}
\begin{center}
\includegraphics[scale=0.34]{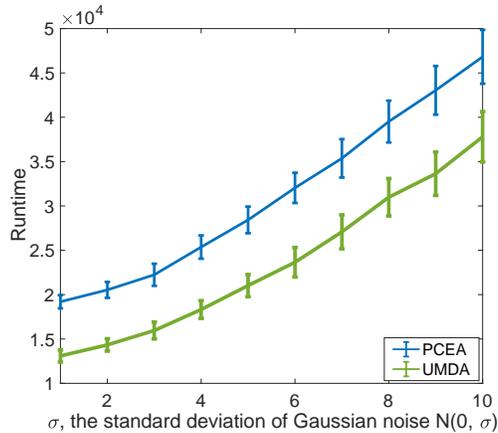}
\caption{Runtime comparison of UMDA and PCEA for noisy \textsc{OneMax}}
\label{fig:onemax_smallnoise_runtime}
\end{center}
\end{figure}

The population size for the PCEA is taken to be $10 \sqrt{n} \log n$ according to the theoretical proofs and empirical study by \cite{PrugelBennett}. According to the proofs by  \cite{dang2015efficient}, the population size $\lambda=\sigma^2 \log{n} $ is chosen for the Mutation-Population algorithm. According to the paper by \cite{Friedrich:2015aa}, the parameter $K=7\sigma^2 \sqrt{n}\log{n}$ is considered for cGA. In presence of additive posterir noise, PBIL and UMDA have not yet been studied much. For  PBIL, the population size is taken as $\lambda = 10n$ (following the theoretical requirement of~\cite{7851024}). From these, we select the best $\mu = \lambda/2$ individuals. In case of UMDA, the total number of generated candidates in a particular generation is chosen as $20 \sqrt{n} \log n$, so that the effective population size is the same as for PCEA. All these parameter settings are retained for all of our experiments in simple and constrained noisy combinatorial optimisation problems.\par

Figure~\ref{fig:onemax_smallnoise}  illustrates a comparison of all of the considered algorithms while solving the noisy \textsc{OneMax} problem for problem size $n=100$. Different levels of Gaussian additive noise with mean 0 and standard deviation $\sigma = 1$ to $10$ are considered in this experiment. It can be seen that PCEA and UMDA are resistant to these noise levels as they are capable of finding the global optimum within the given budget. The runtimes for these two algorithms are shown in Figure~\ref{fig:onemax_smallnoise_runtime}. However, $(1+1)$-EA, Mutation-Population algorithm, PBIL and cGA are not able to cope with even these small levels of noise within the given fixed budget of function evaluations. For these experiments, we run the algorithms until the population converges (which they will, since we do not handle probability margins). The Mann-Whitney U-test is performed on the samples of best results achieved and the runtimes of the algorithms, with the null hypothesis that they are from distributions with equal medians. For each data point, the null hypothesis is rejected at 5\% significance level.  

\subsection{Noisy \textsc{WeightedLinear} problem}

Maximising the \textsc{WeightedLinear} problem as defined above in Section \ref{sec:problem} has only one global optimum, the sum of all the weights. The \textsc{OneMax} problem is a special case of the \textsc{WeightedLinear} problem when all the weights are units. However, optimising the \textsc{WeightedLinear} problem is difficult as the bits with heavier weights get optimised with a higher preference than the bits with lower weights.

\begin{figure}
\begin{center}
\includegraphics[scale=0.38]{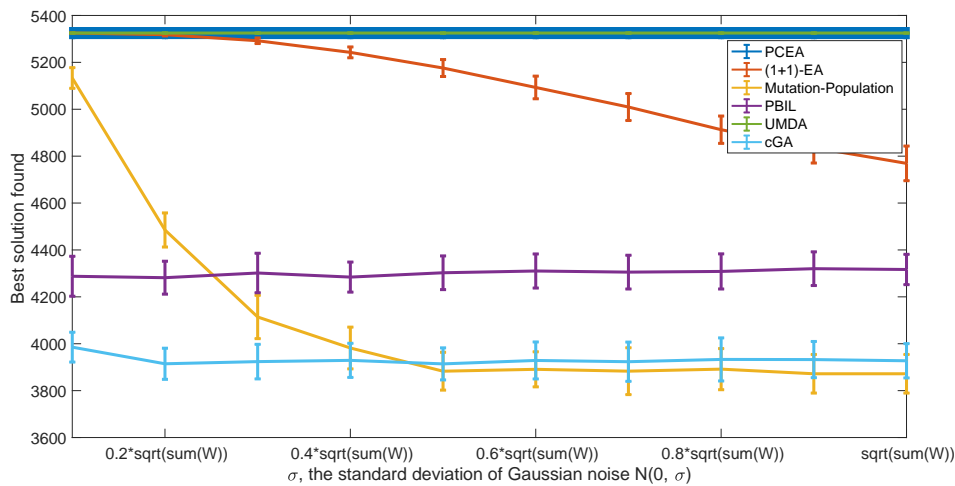}
\caption{Comparison of algorithms while solving noisy \textsc{WeightedLinear}}
\label{fig:linear_smallnoise}
\end{center}
\end{figure}

The plot in Figure~\ref{fig:linear_smallnoise} illustrates the performance comparison of all of the considered algorithms while solving the noisy \textsc{WeightedLinear} problem for the problem size $n=100$. 
Random problem instances were studied with 100 randomly chosen weights between $1$ and $100$. The results for a typical problem is shown in  Figure~\ref{fig:linear_smallnoise} with averages over 100 runs. The standard deviation of the Gaussian noise is shown as multiples of the square root of the sum of the weights. The function evaluation budget allowed to each of the algorithms are fixed at twice the average runtime of PCEA at each noise level (see Table~\ref{tab:linear_budget}).

\begin{figure}
\begin{center}
 \includegraphics[scale=0.3]{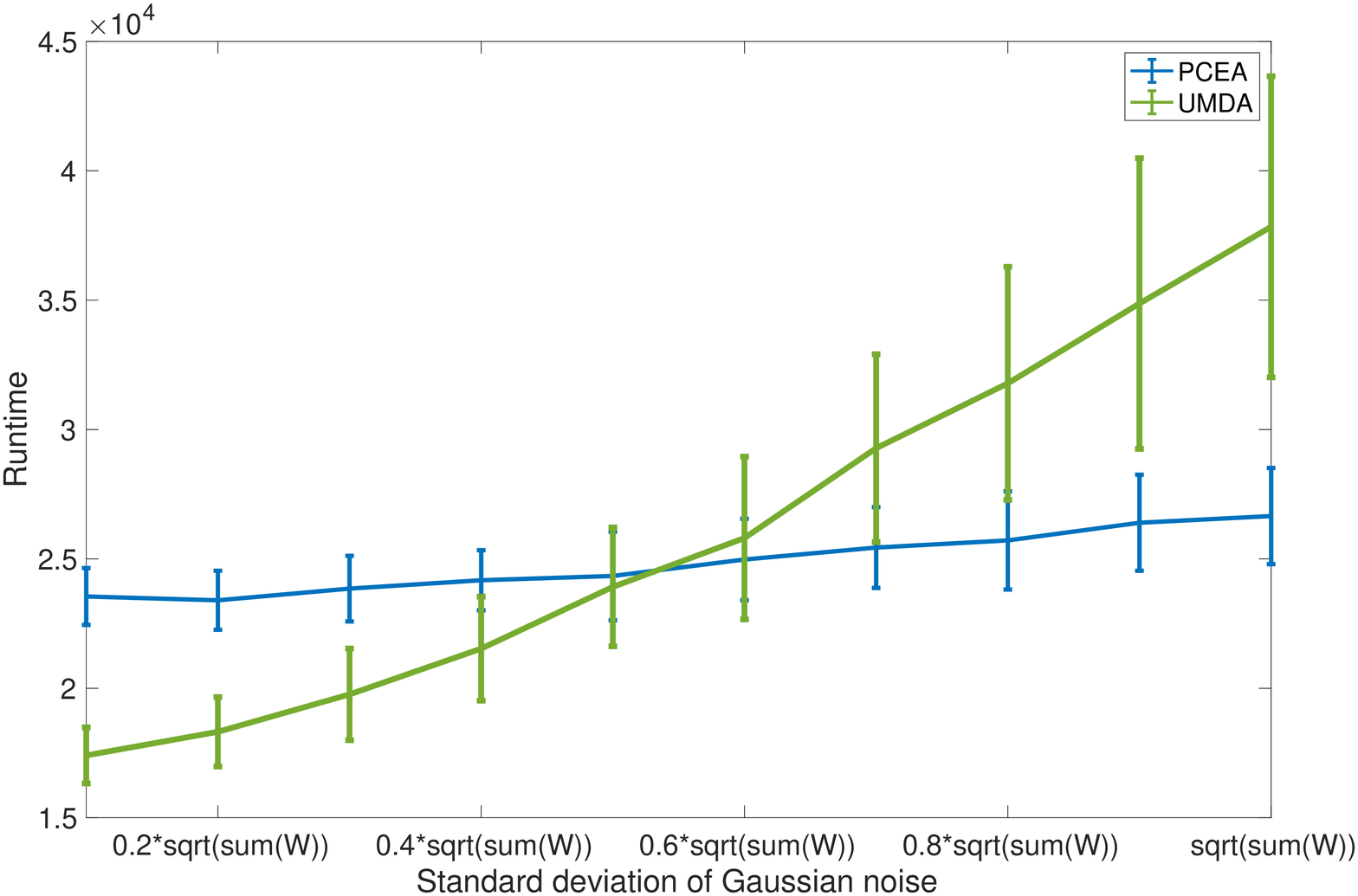}
  \caption{Runtime comparison of UMDA and PCEA for noisy \textsc{WeightedLinear}}
\label{fig:linear_smallnoise_runtime}
\end{center}
\end{figure}

 \begin{table}[H]
  \caption{Function evaluation budgets allowed for noisy \textsc{WeightedLinear} experiments}
    \centering
    \begin{tabular}{|c|c|c|c|c|c|}
    \hline
         $\sigma$ &  1 & 2 & 3 & 4 & 5  \\
         budget & 47096 & 46801 & 47704 & 48350 & 48682 \\
         \hline
         $\sigma$ &  6 & 7 & 8 & 9 & 10 \\
         budget & 49954 & 50876 & 51429 & 52794 & 53310\\
         \hline
     \end{tabular}
    \label{tab:linear_budget}
\end{table}
As evident from Figure~\ref{fig:linear_smallnoise}, the curves of PCEA and UMDA are coincident, showing that they can cope with the noise well and are resistant up to these levels of noise. The runtime of UMDA and PCEA are plotted in Figure ~\ref{fig:linear_smallnoise_runtime}. However, the performance of the $(1+1)$-EA and Mutation-Population algorithm worsen with increasing noise. Even with relatively small noise levels, the cGA and PBIL are not able to solve the problem within twice the runtime of PCEA. 

It is evident from the empirical results of these simple noisy problems that uniform crossover-based PCEA and UMDA can cope with noise significantly better than the other algorithms. At this point, it is interesting to note that, UMDA employs a mechanism similar to \textit{genepool crossover}, where at each bit position, the offspring bit is obtained by recombination of that bit across the whole parent population. It is hypothesised that these two algorithms are therefore highly similar in operation.

\section{Experiments --- Noisy Combinatorial Problems}\label{noisycombinatorialsection}
\subsection{Noisy \textsc{SubsetSum}}
\begin{figure}
\begin{center}
\includegraphics[scale=0.44]{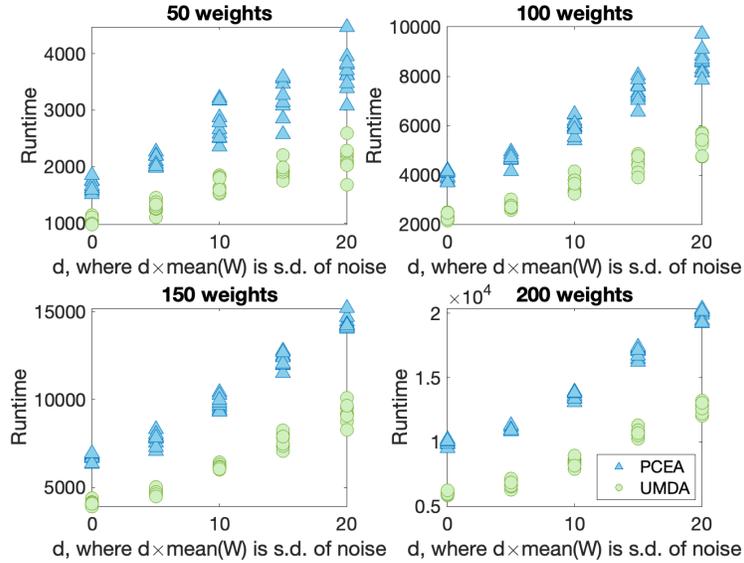}
\caption{Runtime comparison of UMDA (circles) and PCEA (triangles) while solving instances of the noisy \textsc{SubsetSum} problem.}
\label{fig:ssp_runtime}
\end{center}
\end{figure}
 Given the success of UMDA and PCEA on the noisy toy problems, and the failure of the others to cope with even modest levels of noise, we now move to the second stage of the study considering only UMDA and PCEA.

 For the noisy \textsc{SubsetSum} problem, a range of problem sizes is considered with 50, 100, 150 and 200 weights, each lying between 1 and 100, and chosen uniformly at random. Corresponding to each problem size, 10 different problems are considered. The target $\theta$ is considered to be two-third of the sum of all the weights in the set. The additive Gaussian noise considered in the \textsc{SubsetSum} problem is centered at zero and is considered to have standard deviation of integral multiples of the mean of the weights, viz., $5 \times mean(W)$, $10 \times mean(W)$, $15 \times mean(W)$ and $20 \times mean(W)$.\par

The  \textsc{NoisySubsetSum} problem being a minimisation problem, if we obtain the (non-noisy) fitness value of zero, we obtain the global optimum. Both the algorithms are able to find the global optimum for these problems and their corresponding noise levels. We therefore plot the runtime (averaged over 100 runs) to find the optimum against the standard deviation of the noise --- see Figure \ref{fig:ssp_runtime}. Using the Mann-Whitney U-test it is observed that UMDA has the  better runtime.
\subsection{Noisy \textsc{Knapsack} (Version 1)}
 For the first version of the noisy \textsc{Knapsack} problem, instances with 50, 100, 150 and 200 weights (randomly chosen between 1 and 100) with associated profits (in the same range) are considered. The maximum capacity of the knapsack is taken to be two-thirds of the sum of all the weights considered.  \par
 \begin{figure}
\begin{center}
\includegraphics[scale=0.44]{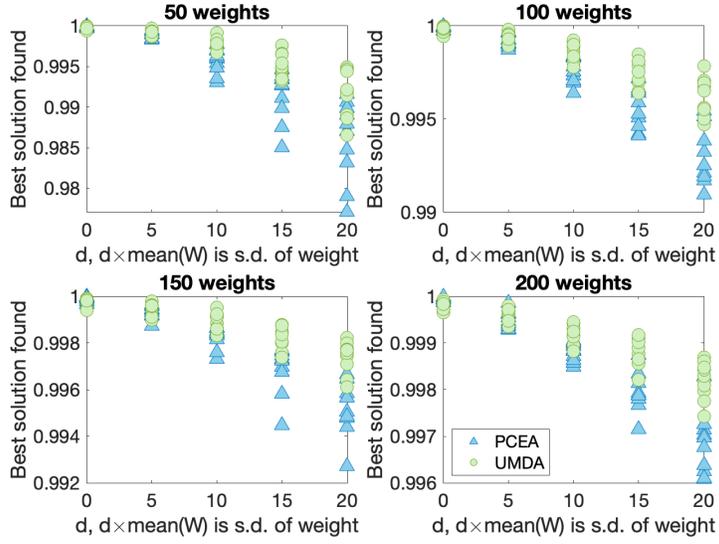}
\caption{Solution quality of UMDA (circles) and PCEA (triangles) while solving instances of \textsc{NoisyKnapsackV1} }
\label{fig:kp1_solution}
\end{center}
\end{figure}
 When noise is added, neither algorithm finds the optimal solution, so we record the best solution found (as assessed by non-noisy fitness function). PCEA is run until the population converges whereas, UMDA is run for twice that time, and we report the time taken to find the best solution encountered. 
 \begin{figure}
\begin{center}
\includegraphics[scale=0.44]{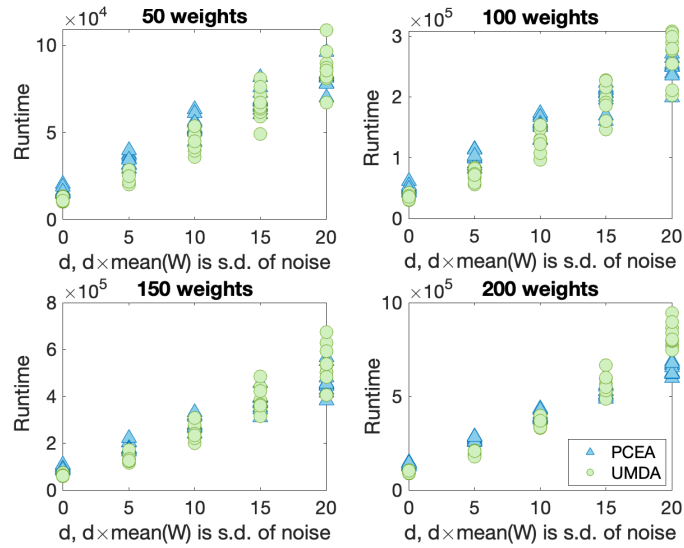}
\caption{Comparison of runtime of UMDA (circles) and PCEA (triangles) while solving the \textsc{NoisyKnapsackV1} }
\label{fig:kp1_runtime}
\end{center}
\end{figure}
 For each problem instance, we plot (in Figure \ref{fig:kp1_solution}) the best solution found (averaged over 100 runs) as a fraction of the best solution ever encountered for that problem instance. This enables us to make meaningful comparisons between problem instances. The best known solution for each problem instance has a scaled fitness value of 1. For each problem size, 10 different problems are considered. Figure \ref{fig:kp1_runtime} shows the time taken (on average) to locate the best found solution in each case. We can observe in Figures \ref{fig:kp1_solution} and \ref{fig:kp1_runtime}, that both the algorithms can find good, though not optimal solutions, for \textsc{NoisyKnapsackV1} with significant levels of noise.  Observations from Mann-Whitney U-test show that UMDA is slightly better than PCEA with these parameter settings.

\subsection{Noisy \textsc{Knapsack} (Version 2)}
\begin{figure}
\begin{center}
\includegraphics[scale=0.42]{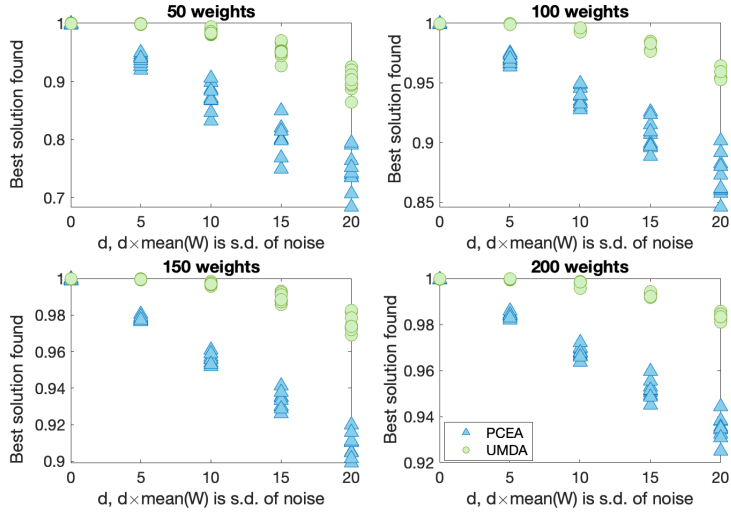}
\caption{Solution quality of UMDA (circles) and PCEA (triangles) while solving the \textsc{NoisyKnapsackV2}}
\label{fig:kp2_solution}
\end{center}
\end{figure}
\begin{figure}
\begin{center}
\includegraphics[scale=0.42]{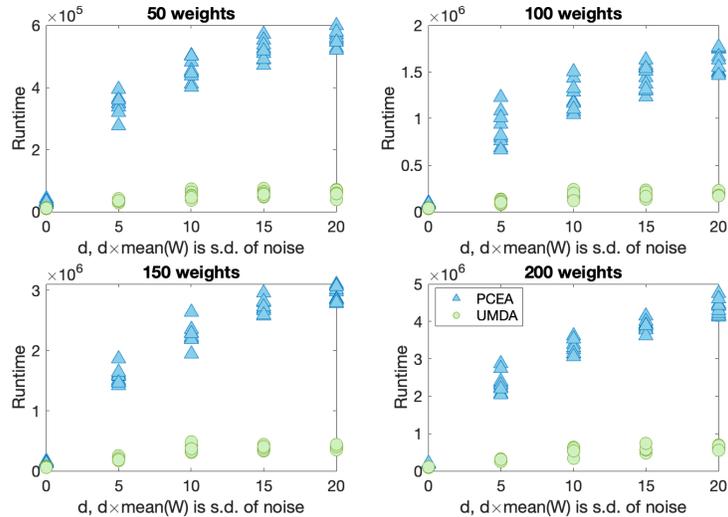}
\caption{Comparison of runtime of UMDA (circles) and PCEA (triangles) while solving the \textsc{NoisyKnapsackV2}}
\label{fig:kp2_runtime}
\end{center}
\end{figure}
When the measurements of the weights is uncertain, as well as the profits, this creates a more complex noise model for the \textsc{Knapsack} problem. In the first stage, the total weight of the proposed solution is compared against the capacity, and this is done with added noise. Hence it may be thought that the proposed solution is feasible when in fact it is not. If it is considered feasible, then the benefit (total profit) is calculated, again with added noise.The parameters are considered as in the previous version of the \textsc{Knapsack} problem. 10 problems each of 50, 100, 150, and 200 weights (lying between 1 and 100) with associated profits (also lying in the same range) are considered.\par

Figure \ref{fig:kp2_solution} depicts how the best (non-noisy) solution varies for different problem sizes. This value is scaled with respect to the best value found when there is no noise. PCEA is run until the population converges while UMDA is run for twice that time, and we report the time taken to find the best solution encountered. The Mann-Whitney U-test shows that the best solution achieved and corresponding runtime of UMDA is better than  PCEA in these particular parameter settings. The runtime required to find these values is shown in Figure \ref{fig:kp2_runtime}, and we see that UMDA finds its best solution considerably faster than PCEA.

\subsection{Noisy \textsc{ConstrainedSetCover} and \textsc{PenaltySetCover}} \label{setcover1}
The \textsc{ConstrainedSetCover} problem is solved by initially finding the feasible solutions and then minimising the number of the selected sets. This lexicographic ordering is achieved in the selection mechanism of the considered algorithms. 

 \begin{figure}
\begin{center}
\includegraphics[scale=0.44]{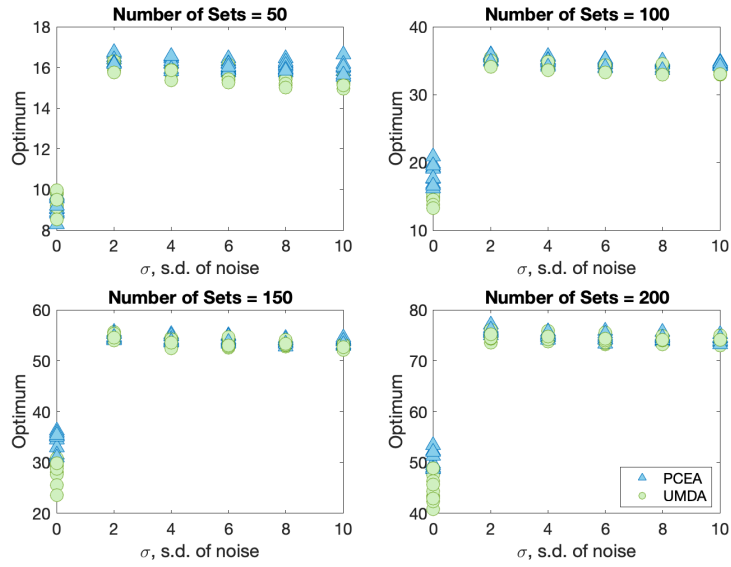}
\caption{Solution quality of UMDA (circles) and PCEA (triangles) while solving the \textsc{ConstrainedSetCover}}
\label{fig:setcover_constrained}
\end{center}
\end{figure}
In PCEA, the child with least uncovered elements is selected. When both of the children have the same number of uncovered elements, the  child with the minimum number of sets goes to the next population. In UMDA, the sorting of the population is based on the above mentioned lexicographic ordering. We consider margin handling in UMDA for all the following experiments in single objective-optimisation.\par

\begin{figure}
\begin{center}
\includegraphics[scale=0.44]{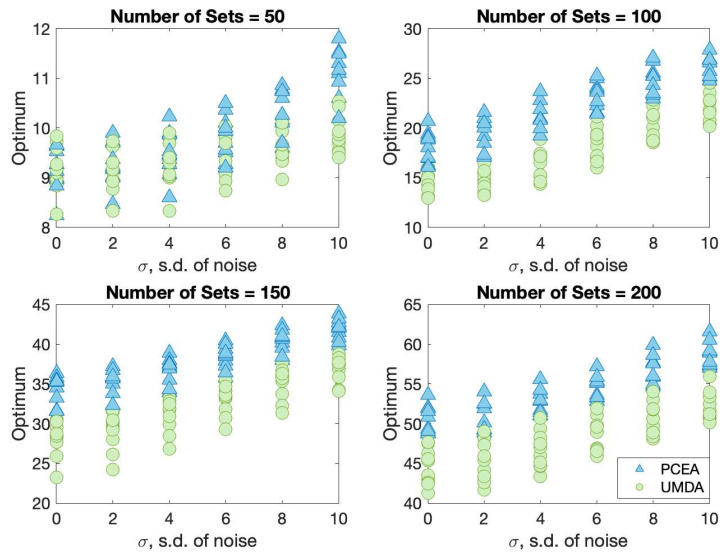}
\caption{Best solution found in stipulated budget of function evaluations by UMDA (circles) and PCEA (triangles) for \textsc{NoisyPenaltySetCover}}
\label{fig:penaltysetcoveropt}
\end{center}
\end{figure}
\begin{figure}[H]
\begin{center}
\includegraphics[scale=0.44]{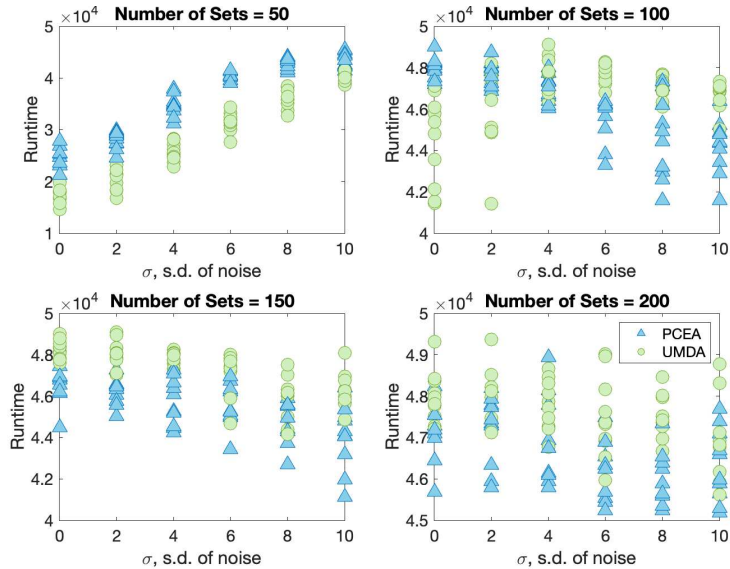}
\caption{Runtime of UMDA (circles) and PCEA (triangles) for best solution found while solving \textsc{NoisyPenaltySetCover}}
\label{fig:penaltysetcoverruntime}
\end{center}
\end{figure}
The alternative \textsc{PenaltySetCover} problem handles the constraint within the penalty function, hence creating a single objective. For both versions of the noisy \textsc{SetCover} problem, a range of 40  problem instances (10 for each problem size) are run with 100 elements and 50, 100, 150 and 200 subsets are available to cover those elements. The problems are created by randomly generating subsets, where the probability of including any element in any subset is $p$. This is set so that the probability of there being cover is large:
\[
(1 - (1-p)^n)^m = 1 - \delta
\]
Therefore, we take:
\[
p = 1 - (1 - (1 - \delta)^{1/m})^{1/n}
\]
We have chosen $\delta=0.001$. All the algorithms are run until 50,000 function evaluations are reached. An average of 30 runs are reported. Figure \ref{fig:setcover_constrained} reports the best feasible solution found in the fixed budget of function evaluations. As evident from the figure, neither of the algorithms can handle noise well. The noisy feasibility check significantly worsens the optimum found even for small standard deviations of noise.

The parameters considered for solving the \textsc{PenaltySetCover} are chosen same as the \textsc{ConstrainedSetCover}.  For each problem, we plot the best feasible solution found so far in the given function evaluation budget and the runtime in Figures \ref{fig:penaltysetcoveropt} and \ref{fig:penaltysetcoverruntime}. It is interesting that both the algorithms can solve the noisy instances in a scalable manner, with UMDA typically producing better quality solutions.

\section{Noisy Combinatorial Multi-Objective Problems}\label{moalgorithms}

In this section, we empirically examine the performances of several evolutionary algorithms on noisy combinatorial multi-objective problems. Much of the previous work on multi-objective optimisation (especially in the context of noise) has concerned continuous problems~(\cite{goh2010investigation,Shim13, Fieldsend15, falcon2020indicator}). In this paper, we focus on discrete problems, but with additive (posterior) Gaussian noise.

A noisy multi-objective combinatorial problem in the search space of binary strings may be defined as follows, 
\begin{equation*} 
    f(\textbf{x}) = (f_1 (\textbf{x})+N(0,\sigma), f_2(\textbf{x})+N(0,\sigma), \dots, f_k(\textbf{x})+N(0,\sigma))
\end{equation*}
where, $\textbf{x} \in \{0, 1\}^n$ is a candidate solution. The objectives $f_1(\textbf{x}),f_2(\textbf{x}), \dots, f_k(\textbf{x})$ are conflicting in nature, so there does not necessarily exist an optimal solution that will minimise all the objectives simultaneously. Instead, there exists a set of non-dominating solutions known as the \textit{Pareto optimal solution set} where none of the objectives may be improved without worsening at least one of the other objectives. In the context of noisy multi-objective optimisation, the goal is to find the set of Pareto optimal solutions, as defined in the absence of noise --- however, the challenge is that each time a comparison is made, noise is applied. This is particularly problematic for algorithms that make use of an \emph{archive} of non-dominated solutions, as it is easy for a solution to be incorrectly placed in the archive due to the noise.
\par
In order to assess how successfully we have approximated the true Pareto optimal set, we measure the spread of a set of non-dominated solutions on the basis of the frequently used \textit{hypervolume} performance indicator (\cite{zitzler1998multiobjective}). Where we seek to minimise each objective, this is a measure of the area (or volume) of the region bounded below by a set of candidate solutions simultaneously and bounded above by a reference point $r$ in the objective space. The reference point $r$ is chosen to be the maximum value each objective function can attain in each corresponding dimension of the objective space, i.e., $r=(\max f_1, \max f_2, \dots, \max f_k)$. Conversely, for maximisation problems, we take the volume between the candidate set and a lower bounding reference point (in the case of non-negative objectives, it is common to take the origin as the reference point). We use hypervolume of the population as an indicator of the spread of the non-dominated solutions in each generation of the considered algorithms.  \par

In this paper, we have studied two noisy multi-objective problems. The first is based on the toy benchmark problem  \textit{Counting Ones Counting Zeroes} (\textsc{COCZ}), in which the first objective function counts the number of ones in a string, and the second objective function counts the number of ones in the first $m$ bits and the number of zeroes in the remainder. We seek to maximise both objectives.
\[
\textsc{NoisyCOCZ}(x) = \]
\[
\Bigg( \sum_{i=1}^n x_i + N(0, \sigma), \sum_{i=1}^m x_i+\sum_{i=m+1}^n(1-x_i)+ N(0, \sigma) \Bigg) \]
The Pareto optimal front consists of strings of the form $1^m *^{(n-m)}$.

The second problem is a multi-objective version of \textsc{SetCover} problem, with the objective function and the constraint as defined in \textsc{ConstrainedSetCover} as the two objective functions. These objectives are conflicting in nature. The first objective minimizes the number of sets required to cover all the $m$ elements of the target set, and the second objective minimizes the number of uncovered elements. The noisy version of the multi-objective \textsc{SetCover} problem is defined as follows,

\[
\textsc{NoisyMulti-objectiveSetCover}(x) = \]
\[\Bigg( \sum_{j=1}^n x_j + N(0, \sigma),
\sum_i [ \sum_{j=1}^n a_{ij}x_j = 0 ] + N(0, \sigma)\Bigg) \]

\section{Algorithms Chosen for Noisy Multi-Objective Combinatorial Problems}

\subsection{Simple Evolutionary Multi-objective Optimiser (SEMO)}

SEMO~(\cite{laumanns2004running}) is one of the simplest evolutionary algorithms designed for multi-objective optimisation in discrete search space. To the best of our knowledge, it has not previously been used to solve noisy problems. SEMO is a simple population-based algorithm using one-bit mutation, and a variable population size (representing the current non-dominated solutions found). The algorithm starts with adding an initial solution $\textbf{x}\in \{0,1\}^n$ chosen uniformly at random to the population $P$. Then a solution $\textbf{y}$ is chosen randomly from $P$ and mutated with a one-bit flip to obtain $\textbf{y'}$. If $\textbf{y'}$ is dominated by anything in $P$ it is discarded. Otherwise it is added to $P$ and all the solutions that $\textbf{y'}$  dominates in $P$ are discarded. Then a new $\textbf{y}$ is chosen from $P$ and the process is repeated. One of the great challenges SEMO will face due to noisy dominance relations is that, often good solutions will be discarded and bad solutions will be retained in $P$.
\begin{algorithm}
Initialise solution $\textbf{x}$ and add to population $P$. \newline
\Repeat{
Choose $\textbf{y}$ from $P$ and mutate a random bit to get $\textbf{y'}$. \\
If $\textbf{y'} $ is not dominated by any solution in P and $\textbf{y'} \not \in P$, add $\textbf{y'}$ to P and discard all solutions in P that $\textbf{y'}$ dominates. 
}
 \caption{SEMO}
 \label{alg:semo}
\end{algorithm}

\subsection{Non-dominated Sorting Genetic Algorithm - II (NSGA-II)}
NSGA-II by \cite{deb2002fast} sorts the population into non-dominated fronts in each generation. Based on non-dominated sorting and using a crowding heuristic to break ties, the best half of individuals become the parent population of the next generation. In case of noisy function evaluations, non-dominated sorting will be affected and worse solutions will appear in better non-dominated fronts. We use the same algorithm structure as defined in ~\cite{deb2002fast} except considering noisy function evaluations during the selection process. 

\subsection {Variants of Multi-objective Univariate Marginal Distribution Algorithm (moUMDA)}
From our experiments in noisy single-objective combinatorial problems, UMDA and PCEA show significantly better performance in handling noise compared to the other algorithms we tried, with UMDA generally producing better quality solutions. From these results, we hypothesise that a multi-objective version of UMDA (denoted moUMDA) may be able to handle large levels of noise in noisy combinatorial multi-objective problems if proper diversification mechanisms are employed. In order to investigate this, we have considered several versions of moUMDA in our analysis with different diversification techniques.

 \cite{pelikan2005multiobjective} introduced a version of UMDA to address multi-objective problems which used non-dominated sorting in the selection mechanism. They also experimented with clustering methods, to help the algorithm generate solutions across the Pareto front. We have followed this idea, and studied several versions of UMDA adapted for multi-objective problems. Where non-dominated sorting and crowding are used for selection, these are implemented identically to NSGA-II. We also consider making use of an archive, and in using hypervolume as a criterion in selection:
 
 \begin{description}
 \item[moUMDA without duplicates] Uses non-dominated sorting (with crowding to break ties) for selection. Maintains diversity by disallowing duplicates when generating the population. See Algorithm~\ref{alg:moUMDA-no-duplicates}.\\
 \item[moUMDA with clustering] Uses non-dominated sorting (with crowding to break ties) for selection. Clusters the selected population members (using either K-means or Hierarchical Agglomeration), and produces a frequency vector for each cluster. Generates next population from these, in proportion to the number of items within each cluster. See Algorithm~\ref{alg:moUMDA-clustering}.\\
 \item[moUMDA with Pareto archive] Maintains an archive of non-dominated solutions and uses this to generate the frequency vector for the next population. Uses non-dominated sorting (with crowding to break ties) for selection, and updates the archive with the selected items. See Algorithm~\ref{alg:moUMDA-archive}.\\
 \item[moUMDA with hypervolume comparison operator] Uses binary tournament selection, comparing solutions initially by Pareto dominance. If neither dominates the other, then select the one with the better hypervolume indicator value. See Algorithm~\ref{alg:moUMDA-hypervolume}.
 \end{description}
 
\begin{algorithm}
Initialise  frequency vector $p=(0.5, \ldots, 0.5)$ \newline
\Repeat{
Generate population of size $\lambda$ from $p$, disallowing duplicates. \\
Use non-dominated sorting and crowding to select the best $\mu$ individuals. \\
Update frequency vector $p$ based on selected individuals.
}
 \caption{moUMDA without duplicates}
 \label{alg:moUMDA-no-duplicates}
\end{algorithm}

\begin{algorithm}
Set $k = \lfloor \sqrt{\mu} \rfloor$ as the number of clusters.\\
Initialise  frequency vectors $p_i=(0.5, \ldots, 0.5)$ for each $i=1 \ldots k$.\\
Set $q_i = \mu/k$ for each $i=1 \ldots k$.\\
\Repeat{
Generate population of size $2 q_i$ from $p_i$, for each $i=1 \ldots k$. \\
Use non-dominated sorting and crowding to select the best $\mu$ individuals from all the populations. \\
Cluster the selected individuals into $k$ clusters. \\
Let $q_i$ be the number of individuals in cluster $i$, for each $i=1 \ldots k$.
Update frequency vectors $p_i$ based on selected individuals in each cluster.
}
 \caption{moUMDA with clustering}
 \label{alg:moUMDA-clustering}
\end{algorithm}

\begin{algorithm}
Initialise  frequency vector $p=(0.5, \ldots, 0.5)$ \newline
Initialise empty archive $P$ \newline
\Repeat{
Generate population of size $\lambda$ from $p$. \newline
Use non-dominated sorting and crowding to select the best $\mu$ individuals. \newline
Add these to archive $P$ and remove any dominated solutions.\newline
Update frequency vector $p$ based on archive $P$.
}
 \caption{moUMDA with Pareto archive}
 \label{alg:moUMDA-archive}
\end{algorithm}

\begin{algorithm}
Initialise  frequency vector $p=(0.5, \ldots, 0.5)$ \\
\Repeat{
Create empty population $P$ \\
\RepeatTimes{$\mu$}{
Generate two strings, $x$ and $y$ from $p$ \\
Add string with best hypervolume to $P$
}
Update frequency vector $p$ based on population $P$.
}
 \caption{moUMDA with hypervolume comparison}
 \label{alg:moUMDA-hypervolume}
\end{algorithm}

 \par

\section{Experiments --- Noisy Multi-objective Problems}
Following the same strategy as for single objective problems, we initially, we choose a wide range of evolutionary multi-objective algorithms to compare their performances on a toy problem:  noisy \textsc{CountingOnesCountingZeroes} (COCZ). The  algorithms considered for solving COCZ consists of SEMO, NSGA-II and several versions of multi-objective UMDA (moUMDA) as described above. Depending on their performances on this problem, we selected a smaller set of the better performing algorithms for the multi-objective noisy \textsc{SetCover} problem. \par

Some recent studies claim that multi-objective evolutionary approaches are useful in solving single objective optimisation problems (\cite{segura2016using}). For example, the multi-objective version of \textsc{SetCover} could enable us to find good solutions to the original single-objective version (by looking at solutions generated which do not violate the constraints). Here, we consider whether this approach is also helpful in the context of noise. 
\par

\subsection{Noisy \textsc{CountingOnesCountingZeroes (COCZ)}}

\begin{figure}[H]
\begin{center}
\includegraphics[scale=0.4]{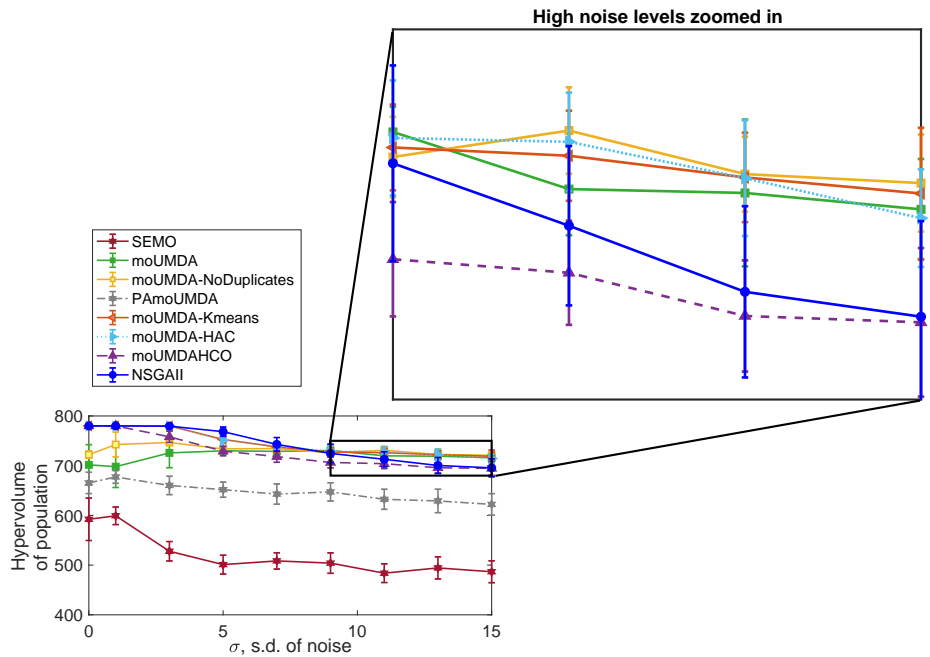}
\caption{Comparison of the hypervolume of population while solving the noisy \textsc{COCZ} with $n=30,m=15$}
\label{fig:cocz_largenoise}
\end{center}
\end{figure}

In this subsection, we solve a toy multi-objective problem, the noisy \textsc{COCZ} with $n=30,m=15$ and with additive Gaussian noise centered at zero and having standard deviations $\sigma=0,1,3,5,7, 9, 11, 13$ and $15$. We set the parameter $\mu=\lambda/2$, where $\lambda=20 \sqrt{n}\log{n}$ for all the versions of moUMDA. For NSGAII, the parent population size is set as $10\sqrt{n}\log{n}$. All the algorithms are run for 50,000 function evaluations and the mean of 30 runs are reported. The best hypervolume of the population found so far in the fixed budget of function evaluations are reported in Figure ~\ref{fig:cocz_largenoise}. 

The Pareto optimal front would contain $2^{15}$ elements and the best possible hypervolume is 780. We have used the dimension-sweep algorithm and the source code by ~\cite{fonseca2006improved} for hypervolume calculation in the experiments.

The results shown in Figure ~\ref{fig:cocz_largenoise} show that SEMO is the worst performing algorithm, even when there is no noise, and the performance degrades slightly as noise is increased.
The Pareto Archive algorithm (PAmoUMDA) is the next worst. Although it does no degrade too much with added noise, it is still clearly worse than the other algorithms. 

The remaining algorithms have rather similar performance, but we can still distinguish different behaviours by looking at the zoomed in section of the plot in Figure ~\ref{fig:cocz_largenoise}.
The version of moUMDA that uses the hypervolume comparison operator (moUMDAHCO) performs very well when there is little or no noise. However, its performance degrades considerably as the level of noise increases. The same is true for NSGAII. When the noise reaches a standard deviation of $\sigma=15$, these two algorithms are the worst of the remaining ones.

The plain moUMDA and the version forbidding duplicates in the population both have the curious property that their performance improves with the presence of low levels of noise, and then degrade at higher levels of noise. We speculate that low levels of noise allow for much more diversity in the populations. At high levels of noise ($\sigma=15$) they are the best performing algorithms, along with the two versions of moUMDA that use clustering (moUMDA-Kmeans and moUMDA-HAC). moUMDA with no duplicates is marginally the best overall at this level of noise.

\subsection{Noisy multi-objective \textsc{SetCover}}
In this section, we compare the performance of three of our multi-objective algorithms, viz., NSGA-II, moUMDA with no duplicates allowed and moUMDA employing K-means clustering, on the  noisy multi-objective \textsc{SetCover} problem. We have chosen these algorithms based on their behaviours on the \textsc{COCZ}. These were amongst the best algorithms we tried on that problem. There being little to distinguish the two different clustering methods, we have chosen to test just one of these (K-means clustering). We have selected the ``no duplicates'' version of moUMDA, as this gave a small advantage over the plain moUMDA. And we have kept NSGAII as this is a standard algorithm for any multi-objective problem.

\begin{figure}[H]
\begin{center}
\includegraphics[scale=0.39]{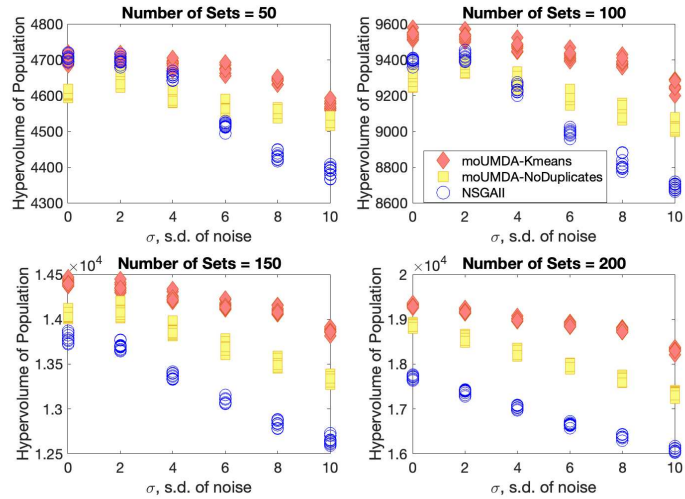}
\caption{Best hypervolume of population obtained for \textsc{Multi-objectiveSetCover}}
\label{fig:mosetcover_runtime}
\end{center}
\end{figure}

\begin{figure}[H]
\begin{center}
\includegraphics[scale=0.39]{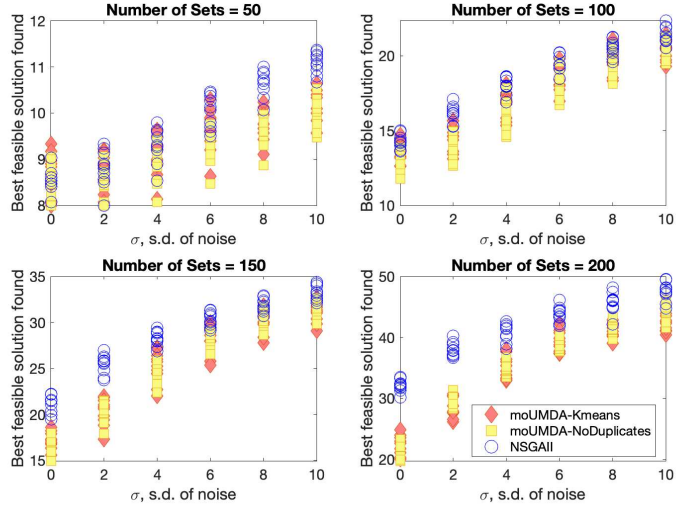}
\caption{Best feasible solution found while solving the noisy \textsc{Multi-objectiveSetCover}}
\label{fig:mosetcover_optimum}
\end{center}
\end{figure}

All the algorithms are run for 50,000 function evaluations.  The best hypervolume of the population obtained in the fixed function evaluation budget for each of 30 runs is shown in Figure~\ref{fig:mosetcover_runtime}. We observe that the clustering algorithm, moUMDA-Kmeans, handles high levels of noise significantly better than other algorithms. It is evident that, the performance of NSGA-II becomes worse as the standard deviation of noise increases and the problem size increases and indeed is the worst of the three algorithms on this problem.

We also consider the multi-objective formulation of noisy \textsc{SetCover} as a means to solving the standard single objective problem. To this end, we consider the quality of the best feasible solutions found by each algorithm, averaged over the 30 runs. The results are plotted in Figure~\ref{fig:mosetcover_optimum}. Again, the two versions of moUMDA perform better than NSGAII. A comparison with Figure~\ref{fig:penaltysetcoveropt} shows that this approach can indeed produce better quality results than the single objective formulation.

\section{Conclusion}
 We have empirically studied a range of evolutionary algorithms on a set of noisy problems. The $(1+1)$-EA, as expected, fails to cope with any degree of posterior noise. Interestingly, some algorithms (the mutation-population algorithm and cGA), where there is a theoretical polynomial runtime for noisy \textsc{OneMax}, fail to be useful in practice compared to some other algorithms. PBIL performs somewhat similar to cGA. The Paired Crossover Evolutionary algorithm handles noise well on both the simple test problems, and on the noisy combinatorial problems we have tried. Interestingly, UMDA also handles these cases well, with even a slightly better performance than PCEA. This may be due to the fact that UMDA has a strong selection method (truncation selection) than PCEA (which uses a tournament on pairs of offspring). Of course, parameter values on each could be tweaked to produce slightly different results -- our key finding is that these are the only algorithms we have tried that seem remotely practical for such problems. It seems likely that UMDA's performance is more due to its relationship with crossover algorithms (such as the genepool crossover), rather than considered as an EDA (such as PBIL). 
 
 We are not aware of any previously published results on noisy combinatorial multi-objective problems. We carefully selected a set of multi-objective algorithms on the basis of the performance on noisy COCZ and tested them on the noisy multi-objective \textsc{SetCover}. We observe that multi-objective UMDA with a simple diversity mechanism that allows no duplicate solutions in the population is effective at solving the noisy \textsc{SetCover} problem in both constrained and multi-objective forms. UMDA can also benefit from using a clustering approach when dealing with noisy multi-objective problems.

\small

\bibliographystyle{apalike}
\bibliography{sample}

\end{document}